\documentclass[journal]{IEEEtran}
\usepackage{algorithm}
\usepackage{algorithmicx}
\usepackage{algpseudocode}
\usepackage{amsmath}
\usepackage{mathrsfs}
\usepackage{amsfonts}
\usepackage[numbers,sort&compress]{natbib}
\usepackage{amssymb}
\usepackage{bm}
\usepackage{enumerate}
\usepackage[pdftex]{graphicx}
\usepackage{lineno}
\emergencystretch=\maxdimen
\ifCLASSINFOpdf
\else
\fi

\begin{document}
%
\title{Learning a Virtual Codec Based on Deep \\Convolutional Neural Network to Compress Image}
%
%
%

\author{Lijun~Zhao,
        Huihui~Bai,~\IEEEmembership{Member,~IEEE,}
        Anhong~Wang,~\IEEEmembership{Member,~IEEE,}
        and~Yao~Zhao,~\IEEEmembership{Senior~Member,~IEEE}

\thanks{L.  Zhao, H. Bai, Y. Zhao are with the Beijing Key Laboratory of Advanced Information Science and Network Technology, Institute Information Science, Beijing Jiaotong University, Beijing, 100044, P. R. China, e-mail: {15112084, hhbai, yzhao}@bjtu.edu.cn.}
\thanks{A. Wang is with Institute of Digital Media \& Communication, Taiyuan University of Science and Technology, Taiyuan, 030024, P. R. China, e-mail:wah\_ty@163.com}
}
%
%

\markboth{Journal of \LaTeX\ Class Files}
{Shell \MakeLowercase{\textit{et al.}}: Bare Demo of IEEEtran.cls for IEEE Journals}
%



\maketitle

\begin{abstract}
Although deep convolutional neural network has been proved to efficiently eliminate coding artifacts caused by the coarse quantization of traditional codec, it's difficult to train any neural network in front of the encoder for gradient's back-propagation. In this paper, we propose an end-to-end image compression framework based on convolutional neural network to resolve the problem of non-differentiability of the quantization function in the standard codec. First, the feature description neural network is used to get a valid description in the low-dimension space with respect to the ground-truth image so that the amount of image data is greatly reduced for storage or transmission. After image's valid description, standard image codec such as JPEG is leveraged to further compress image, which leads to image's great distortion and compression artifacts, especially blocking artifacts, detail missing, blurring, and ringing artifacts. Then, we use a post-processing neural network to remove these artifacts. Due to the challenge of directly learning a non-linear function for a standard codec based on convolutional neural network, we propose to learn a virtual codec neural network to approximate the projection from the valid description image to the post-processed compressed image, so that the gradient could be efficiently back-propagated from the post-processing neural network to the feature description neural network during training. Meanwhile, an advanced learning algorithm is proposed to train our deep neural networks for compression. Obviously, the priority of the proposed method is compatible with standard existing codecs and our learning strategy can be easily extended into these codecs based on convolutional neural network. Experimental results have demonstrated the advances of the proposed method as compared to several state-of-the-art approaches, especially at very low bit-rate.
\end{abstract}


\IEEEpeerreviewmaketitle
\section{Introduction}
%
%
%
%
\IEEEPARstart{I}{mage} and video compression is an essential and efficient tool to reduce the amount of social media data and multimedia data on the Internet. Traditional image compression standards such as JPEG, and HEVC, etc., are built on block-wise transformation and quantization coding framework, which can largely reduce image block's redundancy \cite{r1, rrr1, rr1}. However, the quantization after individual block transformation inevitably results in the blocking artifacts during image coding. Meanwhile, large quantization parameters are always assigned to the codec in order to achieve low bit-rate coding leading to serious blurring and ringing artifacts \cite{r3,r4,r5}, when the transmission band-width is very limited. In order to alleviate the problem of Internet transmission congestion \cite{rr2}, advanced coding techniques, such as de-blocking, and post-processing \cite{rrr2}, are still the hot and open issues to be researched.

The post-processing technique for compressed image can be explicitly embedded into the codec to improve the coding efficiency and reduce the artifacts caused by the coarse quantization. For instance, adaptive de-blocking filtering is designed as a loop filter and integrated into H.264/MPEG-4 AVC video coding standard \cite{r7}, which does not require an extra frame buffer at the decoder. The advantage for de-blocking filtering inside codec is to make sure that an established level of image quality is coded and conveyed in the transmission channel. However, the drawback caused by this kind of filtering is the relatively high computational complexity. In order to avoid this drawback and make filtering compatible to traditional codec, the alternative flexible way is to use the filtering as a post-processing operation after image decoding. In \cite{r6}, two methods are introduced to reduce the blocking effects: filtering method and overlapped block based method.

To date, a large amount of methods have been studied to remove compression artifacts by efficient filtering or other algorithms. In \cite{r8}, a wavelet-based algorithm uses three-scale over-complete wavelet to de-block via a theoretical analysis of blocking artifacts. In \cite{r9}, through image's total variation analysis to define two kinds of regions, adaptive bilateral filters is used as an image de-blocking method to differentially deal with these two regions. In \cite{r10}, by defining a new metric involving smoothness of the block boundaries and image content's fidelity to evaluate the blocking artifacts, quantization noise on the blocks is removed by non-local means filter. For image's de-noising and de-blocking, both hard-thresholding and empirical Wiener filtering are carried on the shape adaptive discrete cosine transform domain (DCT), where the support of an arbitrarily shaped transform is adaptively calculated for all the pixels \cite{r11}.

Except the above mentioned methods \cite{r8, r9, r10, r11}, many works have incorporated some priors or expert knowledge into their model. In \cite{r13}, the compression artifacts are reduced by adaptively estimating DCT coefficients in overlapped transform-blocks and integrating the quantization noise model with block similarity prior model. In \cite{r14}, maximum a posteriori criterion is used to resolve the problem of compressed image's post-processing by treating post-processing as an inverse problem. In \cite{r15}, an artifact reducing approach is developed to reduce the artifacts of JPEG compression by dictionary learning and total variation regularization. In \cite{r18}, by using constrained non-convex low-rank model, image de-blocking is formulated as an optimization problem within maximum a posteriori framework for image de-blocking. In \cite{r19}, by the combination of both JPEG prior knowledge and sparse coding expertise, deep dual-domain based restoration is developed for JPEG-compressed images. In \cite{r20}, by exploiting the redundancy of residual in the JPEG streams and the properties of sparsity in the latent images, compressed image's restoration is regarded as a sparse coding process carried out jointly in the DCT and pixel domains. Different from \cite{r13, r14, r15, r18, r19, r20}, the technique of structure-texture decomposition has been used in \cite{r16} to reduce the compression artifacts for JPEG compressed images as well as image contrast enhancement.

Unlike the specific task of compression artifact removal, image de-noising is a more general technique to remove the noise such as additive Gaussian noise, equipment noise, compression artifact and so on. In \cite{r12}, based on a sparse representation on transform domain, an advanced image de-noising strategy is used to achieve collaborative filtering by the following steps: grouping similar 2-D image fragments into 3-D data arrays, 3-D transformation of a group, shrinkage of the transformation spectrum, and inverse 3-D transformation. In \cite{r17}, by exploiting image's nonlocal self-similarity, weighted nuclear norm minimization problem is studied for image de-noising, while the solutions of this problem are well analyzed under different weighting conditions. In \cite{rrr3}, self-learning based image decomposition is applied for single image denoising, while an over-complete dictionary is learned from input image's high spatial frequency for image's reconstruction.

Deep learning has achieved great success for the high-level tasks in the field of computer vision \cite{r36}, such as image classification, object detection and tracking, and image semantic segmentation, etc. Meanwhile, it has made a milestone result again and again for low-level image processing, such as de-noising, image super-resolution, and image in-painting. In the early time, a plain multi-layer perceptron has been employed to directly learn a projection from a noisy image to a noise-free image \cite{r21}. Recently, by directly learning an end-to-end mapping from the low-resolution image to high-resolution one, convolutional neural network is used to resolve the problem of image super-resolution \cite{r22}. More importantly, a general solution to image-to-image translation problems is got with the conditional generative adversarial networks \cite{r24}. Latter, conditional generative adversarial network is used to resolve the problem of multi-tasks learning in \cite{r23} such as: color image super-resolution and depth image super-resolution problems at the same time, and simultaneous image smoothing and edge detection.

From the literatures of \cite{r36, r21, r22, r23, r24}, it can be found that deep learning has been widely applied into various fields. For compression artifact suppression in JPEG compressed images, some literatures have pioneered to eliminate compression artifacts with deep convolutional neural networks. In \cite{r26}, a 12-layer deep convolutional neural network with hierarchical skip connections is used to be trained with a multi-scale loss function. In \cite{r25}, a conditional generative adversarial framework is trained by replacing full size patch generation with sub-patch discrimination to remove compression artifacts and make the enhanced image to be looked very realistic as much as possible. Besides, several works \cite{r28,r29,r30} have used convolutional neural network to compress image, which have got appealing results and make a great preparation and contribution for next image compression standard. However, existing standard codecs are still widely used all over the world, so how convolutional neural network based coding is compatible to the traditional codec remains an open issue.

Recently, two convolutional neural networks are trained using a unified optimization method to cooperate each other so as to get a compact intermediate for encoding and to reconstruct the decoded image with high quality \cite{r27}. Although a unified end-to-end learning algorithm is presented to simultaneously learn these two convolutional neural networks, its approximation by directly connecting two convolutional neural networks before and after codec is not optimal. Our intuitive idea of getting the optimal solution for this problem is to use the convolutional neural network to perfectly replace the classic codec for gradient back-propagation. However, this task is still a challenging work now. Although image-to-image translation \cite{r23,r24} achieves many visually pleasant results which look very realistic, it is difficult to learn a mapping from input image to decoded image for standard codec. Fortunately, the projection from the valid description image to the post-processed compressed image can be well learned by convolutional neural network.

\begin{figure*}[ht]
\centering
\includegraphics[width=6.5in]{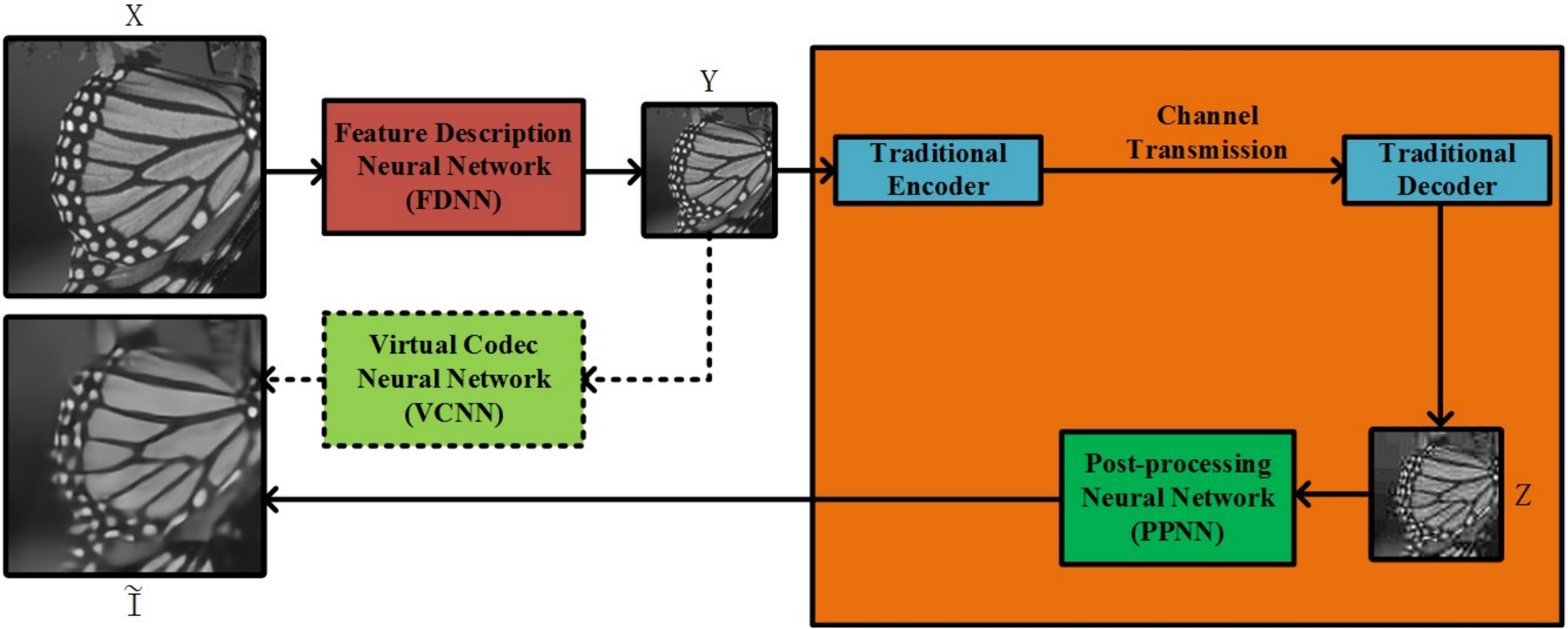}
\caption{The framework of learning a virtual codec neural network to compress image}
\label{Fig1}
\end{figure*}

In this paper, we propose a new end-to-end neural network framework to compress image by learning a virtual codec neural network (denoted as VCNN). Firstly, the feature description neural network (FDNN) is used to get a valid description in the low-dimension space with respect to the ground-truth image, so the amount of image data can be greatly reduced by the FDNN network. After image's valid description, standard image codec such as JPEG is leveraged to further compress image, which leads to image's great distortion and compression artifacts. Finally, we use a post-processing neural network (PPNN) to remove these compression artifacts. The experimental results will validate the efficiency of the proposed method, especially in the case of very low bit-rate. Our contributions are listed as follows:
\begin{enumerate}
  \item In order to efficiently back-propagate the gradient from PPNN network to FDNN network during training, VCNN network is proposed to get an optimal approximation for the projection from the valid feature description image to the post-processed compressed image.
  \item Due to the difficulty of directly training the whole framework once, the learning of three convolutional neural networks in our framework can be decomposed into three sub-problems learning. Although three convolutional neural networks are used during the training, only two convolutional neural networks are used for testing.
  \item Apparently, our framework is compatible with standard codec, so there is no need to change any part in the standard codec. Meanwhile, our learning strategy can be easily extended into these codecs based on convolutional neural network.
\end{enumerate}
The rest of this paper is arranged as follows. Firstly, we give a detail description about the proposed method in Section 2, which is followed by the experimental results in the Section 3. At last, we give a conclusion in the Section 4.

\section{The methodology}

In this paper, we propose a novel way to resolve the problem of non-differentiability of quantizaion function after block transformation in the classic codec, e.g., JPEG, when both convolutional neural networks and traditional codec are used to compress image at very low bit-rate. This way is to learn a virtual codec neural network to optimally approximate the mapping from feature description image to post-processed compressed image.

Our framework is composed of a standard codec (e.g., JPEG), and three convolutional neural networks: FDNN network, PPNN network, and VCNN network, as shown in Fig. \ref{Fig1}. In order to greatly reduce the amount of image data for storage or transmission, we use the FDNN network to get a valid description of $\bm{Y}$ in the low-dimension space with respect to the ground-truth image $\bm{X}$ with size of $M \cdot N$ before image compression. For simplicity, the FDNN network is expressed as a non-linear function $f(\bm{X},\alpha)$, in which $\alpha$ is the parameter set of FDNN network. The compression procedure of standard codec is described as a mapping function $\bm{Z}=g(\bm{Y},\beta)$, where $\beta$ is the parameter set of codec. Our PPNN network learns a post-processing function $h(\bm{Z},\gamma)$ from image $\bm{Z}$ to image $\bm{X}$ to remove the noise, such as blocking artifacts, ringing artifacts and blurring, which are caused by coarse quantization after the separate blocking transformation. Here, the parameter $\gamma$ is the parameter set of PPNN network.

In order to combine the standard codec with convolutional neural network for compression, the direct way is to learn a neural network to approximate the compression procedure of codec. Although convolutional neural network is a powerful tool to approximate any nonlinear function, it's well-known that it's hard to imitate the procedure of image compression. This reason is that the quantization operator is conducted separately on the transformation domain in each block after the DCT transform, which leads to serious block artifacts and coding distortion. However, as compared to the compressed images $\bm{Z}$, the post-processed compressed image $\bm{\tilde{I}}$ has less distortion, because $\bm{\tilde{I}}$ loses some detail information, but does not have obvious artifacts and blocking artifacts. Therefore, the function $h(g(\bm{Y},\beta),\gamma)$ of two successive procedure of codec $g(\bm{Y},\beta)$ and post-processing $h(\bm{Z},\gamma)$ can be well represented by the VCNN network. To make sure that the gradient can be rightly back-propagated from the PPNN to FDNN, our VCNN network is proposed to learn a projection function $v(\bm{Y},\theta)$ from valid feature description $\bm{Y}$ to final output $\bm{\tilde{I}}$ of PPNN. Here, the parameter $\theta$ is the parameter set of VCNN network. This projection can properly approximate the two successive procedure: the compression of standard codec and post-processing based on convolutional neural network. After training the VCNN network, we can use this network to supervise the training of our FDNN network.
\begin{figure*}[ht]
\centering
\includegraphics[width=6.3in]{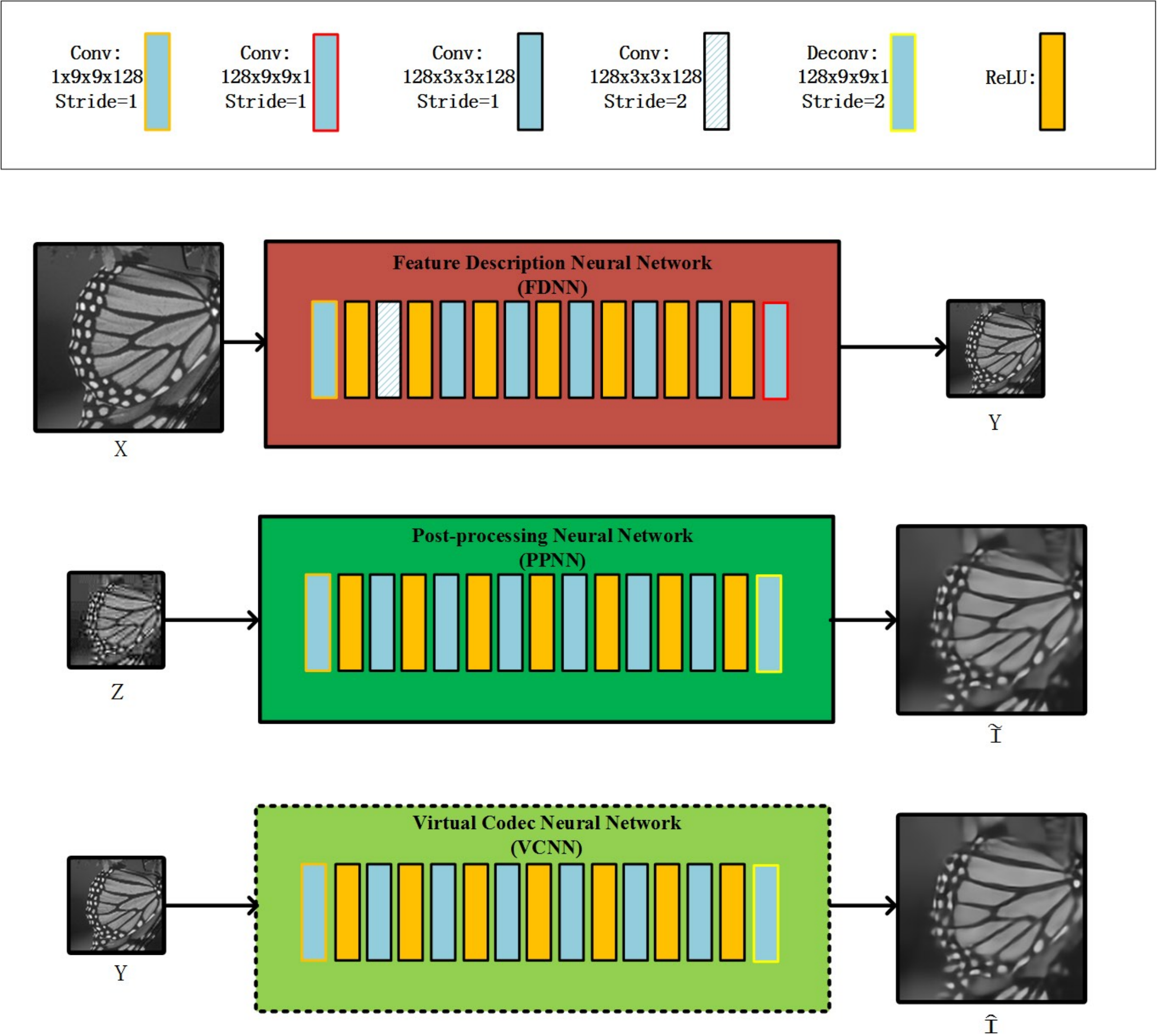}
\caption{The structure of the proposed three convolutional neural networks: FDNN, PPNN, and VCNN}
\label{Fig2}
\end{figure*}

\subsection{Objective function}
Our objective function is written as follows:
\begin{equation}
\begin{split}
\mathop{\arg\min}_{\alpha, \gamma, \theta} L(\bm{X},\bm{\tilde{I}})+ L(\bm{\hat{I}},\bm{\tilde{I}})+L_{SSIM}(s(\bm{Y}),\bm{X}), \\
\bm{Y}=f(\bm{X},\alpha),\bm{\tilde{I}}=h(\bm{Z},\gamma), \bm{Z}=g(\bm{Y},\beta), \bm{\hat{I}}=v(\bm{Y},\theta),
\end{split}
\end{equation}
where $\alpha$, $\gamma$, and $\theta$ are respectively three parameter sets of FDNN, PPNN, and VCNN, and $s(\cdot)$ is the linear up-sampling operator so that $s(\bm{Y})$ and $\bm{X}$ could have the same image size. Here, in order to make final output image $\bm{\tilde{I}}$ to be similar to $\bm{X}$, $L(\bm{X},\bm{\tilde{I}})$ have the L1 content loss $L_{content}(\bm{X},\bm{\tilde{I}})$ and L1 gradient difference loss $L_{gradient}(\bm{X},\bm{\tilde{I}})$ for the regularization of training the FDNN network:
\begin{equation}
\begin{split}
L_{content}(\bm{X},\bm{\tilde{I}})= \frac{1}{M \cdot N}\sum_{i}(||\bm{X}_i-\bm{\tilde{I}}_i||_1),
\end{split}
 \label{eqn::data loss}
\end{equation}

\begin{equation}
\begin{split}
L_{gradient}(\bm{X},\bm{\tilde{I}})= \frac{1}{M \cdot N} \sum_{i} ((\sum_{k\in{\Omega}}||\nabla_k \bm{X}_i)-\nabla_k \bm{\tilde{I}}_i||_1))
\end{split}
\label{eqn::GD loss}
\end{equation}
where $||\cdot||_1$ is the L1 norm, which has better performance to supervise convolutional neural network's training than the L2 norm. This has been reported in the literature of \cite{r33}, which successfully learns to predict future images from the video sequences.

Since standard codec, as a big obstacle, exists between PPNN network and FDNN network, it's tough to make the gradient back-propagate between them. Therefore, it's a challenging task to train FDNN network directly without the supervision of PPNN network. To address this task, we can learn a nonlinear function from the $\bm{Y}$ to $\bm{\tilde{I}}$ in the VCNN network, where the L1 content loss $L_{content}(\bm{\hat{I}},\bm{\tilde{I}})$  and L1 gradient difference loss $L_{gradient}(\bm{\hat{I}},\bm{\tilde{I}})$ in Eq. (4-5) are used to supervise the VCNN network's training. Here, $\bm{\hat{I}}$ is the result predicted by VCNN network to approximate $\bm{\tilde{I}}$.

\begin{equation}
\begin{split}
L_{content}(\bm{\hat{I}},\bm{\tilde{I}})= \frac{1}{M \cdot N}\sum_{i}(||\bm{\hat{I}}_i-\bm{\tilde{I}}_i||_1)
\end{split}
 \label{eqn::CONTENT}
\end{equation}

\begin{equation}
\begin{split}
L_{gradient}(\bm{\hat{I}},\bm{\tilde{I}})= \frac{1}{M \cdot N} \sum_{i} ((\sum_{k\in{\Omega}}||\nabla_k \bm{\hat{I}}_i)-\nabla_k \bm{\tilde{I}}_i||_1))
\end{split}
\label{eqn::GRADIENT}
\end{equation}
Moreover, we hope that feature description image's structural information is similar to ground-truth image $\bm{X}$, so the SSIM loss $L_{SSIM}(s(\bm{Y}),\bm{X})$ \cite{r32, r25} is used to further supervise the learning of FDNN, except the loss from the network of VCNN, which is defined as follows:
\begin{equation}
\begin{split}
L_{SSIM}(s(\bm{Y}),\bm{X})=-\frac{1}{M \cdot N} \sum_{i} L_{SSIM}(s(\bm{Y})_i,\bm{X}_i)
\end{split}
\label{eqn::SSIMLOSS}
\end{equation}

\begin{align}
&L_{SSIM}(s(\bm{Y})_i,\bm{X}_i)=\notag\\
&\frac{(2\mu_{s(\bm{Y})_i}\cdot \mu_{\bm{X}_i}+c1)(2\sigma_{s(\bm{Y})_i \bm{X}_i}+c2)}{(\mu^2_{s(\bm{Y})_i}+\mu^2_{\bm{X}_i}+c1)(\sigma^2_{s(\bm{Y})_i}+\sigma^2_{\bm{X}_i}+c2)}
\end{align}
where $c1$ and $c2$ are two constant values, which respectively equal to $0.0001$ and $0.0009$. $\mu_{\bm{X}_i}$ and $\sigma^2_{\bm{X}_i}$ respectively denote the mean value and the variance of the neighborhood window centered by pixel $i$ in the image $\bm{X}$. In this way, $\mu_{s(\bm{Y})_i}$ as well as $\sigma^2_{s(\bm{Y})_i}$ can be denoted similarly. Meanwhile, $\sigma_{s(\bm{Y})_i \bm{X}_i}$ is the covariance between neighbourhood windows centered by pixel $i$ in the image $\bm{X}$ and in the image $s(\bm{Y})$. Because the function of SSIM is differentiable, the gradient can be efficiently back-propagated during the FDNN network's training.

\subsection{Proposed Network}
As depicted in Fig. \ref{Fig2}, eight convolutional layers in the FDNN network are used to extract features from the ground-truth image $X$ to get a valid feature description $Y$, whose weights of convolutional layer are in the spatial size of 9x9 for the first layer and the last layer, which could make receptive field (RF) of convolutional neural networks to be large enough. In addition, other six convolutional layers in the FDNN use 3x3 convolution kernel to further enlarge the size of RF. In this figure, "Conv:1x9x9x128" denotes the convolutional layer, where the channel number of the input image is 1, the convolution kernel is 3x3 in the spatial domain, and the number of output feature map is 128. Meanwhile, other convolutional layers can be marked similarly. These convolutional layers are used to increase the nonlinearity of the network, when ReLU is followed to activate the output features of these convolutional hidden layers. The feature map number of 1-7 convolutional layers is 128, but the last layer only has one feature map so as to keep consistent with the ground truth image $\bm{X}$. Each convolution layer is operated with a stride of 1, except that the second layer uses stride step of 2 to down-sample feature maps, so that the convolution operation is carried out in the low-resolution space to reduce computational complexity from the third convolutional layer to the 8-th convolutional layer. All the convolutional layers are followed by an activation layer with ReLU function, except the last convolutional layer.

In the PPNN network, as shown in Fig. \ref{Fig2}, we leverage seven convolutional layers to extract features and each layer is activated by ReLU function. The size of convolutional layer is 9x9 in the first layer and the left six layers use 3x3, while the output channel of feature map equals to 128 in these convolutional layer. After these layers, one de-convolution layer with size of 9x9 and stride to be 2 is used to up-scale feature map from low-resolution to high-resolution so that the size of output image is matched with the ground truth image.

We design the VCNN network to be the same structure with the PPNN network, as displayed in Fig. \ref{Fig2}, because they belong to the same class of low-level image processing problems. From Fig. \ref{Fig2}, it also can be found that the VCNN network works to make the valid feature description image $\bm{Y}$ degrade to a post-processed compressed but high-resolution image $\bm{\tilde{I}}$. On the contrary, the functionality of the PPNN network is to improve the quality of the compressed feature description image $\bm{Z}$ so that the user could receive a high-quality image $\bm{\tilde{I}}$ without blocking artifacts and ringing artifacts after post-processing with PPNN network at the decoder.
\begin{algorithm}[!h]
\caption{Learning Algorithm for Training Our Three Convolutional Neural Networks: FDNN, PPNN, and VCNN}
\scriptsize
\begin{algorithmic}[1]
\renewcommand{\algorithmicrequire}{\textbf{Input:}}
\renewcommand{\algorithmicensure}{\textbf{Output:}}
\Require Ground truth image: $\bm{X}$; the number of iteration: $K$; the total number of images for training: $n$; the batch size during training: $m$;
\Ensure  The parameter sets of FDNN network and PPNN network: $\alpha$, $\gamma$;
\State The initialization of the FDNN network's output by down-sampling to prepare for the training of PPNN network;
\State The initialization of parameter sets: $\alpha$, $\beta$, $\gamma$, $\theta$;
\For{$k=1$ to $K$}
    \State The valid description images are compressed by standard codec with $\beta$
    \For{$epoch=1$ to $p$}
        \For{$i=1$ to floor$(n/m)$}
            \State Update the parameter set of $\gamma$ by training the PPNN network to minimize
            \State the Eq. (2-3) with $i$-th batch images
        \EndFor
    \EndFor
    \For{$epoch=1$ to $p$}
        \For{$j=1$ to floor$(n/m)$}
            \State Update the parameter set of $\theta$ by training the VCNN network to minimize
            \State the Eq. (4-5) with $j$-th batch images
        \EndFor
    \EndFor
    \For{$epoch=1$ to $q$}
        \For{$l=1$ to floor$(n/m)$}
            \State Update the parameter set of $\alpha$ with fixing $\theta$ by training the FDNN
            \State network to minimize Eq. (2-3) and Eq. (6-7) with $l$-th batch images
        \EndFor
    \EndFor
\EndFor
\State Update the parameter set of $\gamma$ by training the PPNN network to minimize the Eq. (2-3)
\State \textbf{return} $\alpha$, $\gamma$;
\end{algorithmic}
\end{algorithm}

\subsection{Learning Algorithm}
Due to the difficulty of directly training the whole framework once, we decompose the learning of three convolutional neural networks in our framework as three sub-problems learning. First, we initialize all the parameter set $\beta$, $\alpha$, $\gamma$, and $\theta$ of codec, FDNN network, PPNN network, and VCNN network. Meanwhile, we uses Bicubic, Nearest, Linear, Area, and LANCZOS4 interpolation methods to get an initial feature description image $\bm{Y}$ of the ground-truth image $\bm{X}$, which is then compressed by JPEG codec as the input of training data set at the beginning. Next, the first sub-problem learning is to train PPNN network by updating the parameter set of $\gamma$ according to the Eq. (2-3). The compressed description image $\bm{Z}$ got from ground-truth image $\bm{X}$ and its post-processed compressed image $\bm{\tilde{I}}$ predicted by PPNN network are used for the second sub-problem's learning of VCNN to update parameter set of $\theta$ based on the Eq. (4-5). After VCNN's learning, we fix the parameter set of $\theta$ in the VCNN network to carry on the third sub-problem learning by updating the parameter set of $\alpha$ for training FDNN network according to Eq. (2-3) and Eq. (6-7). After FDNN network's learning, the next iteration begins to train the PPNN network, after the updated description image are compressed by the standard codec. Our whole training process is summarized in the \textbf{Algorithm-1}. It is worth mentioning that the functionality of VCNN network is to bridge the great gap between FDNN and PPNN. Thus, once the training of our whole framework is finished, the VCNN network is not in use any more, that is to say, only the parameter sets of $\alpha$, $\gamma$ in the networks of FDNN and PPNN are used during testing.

\begin{figure}[!ht]
\centering
\includegraphics[width=3in]{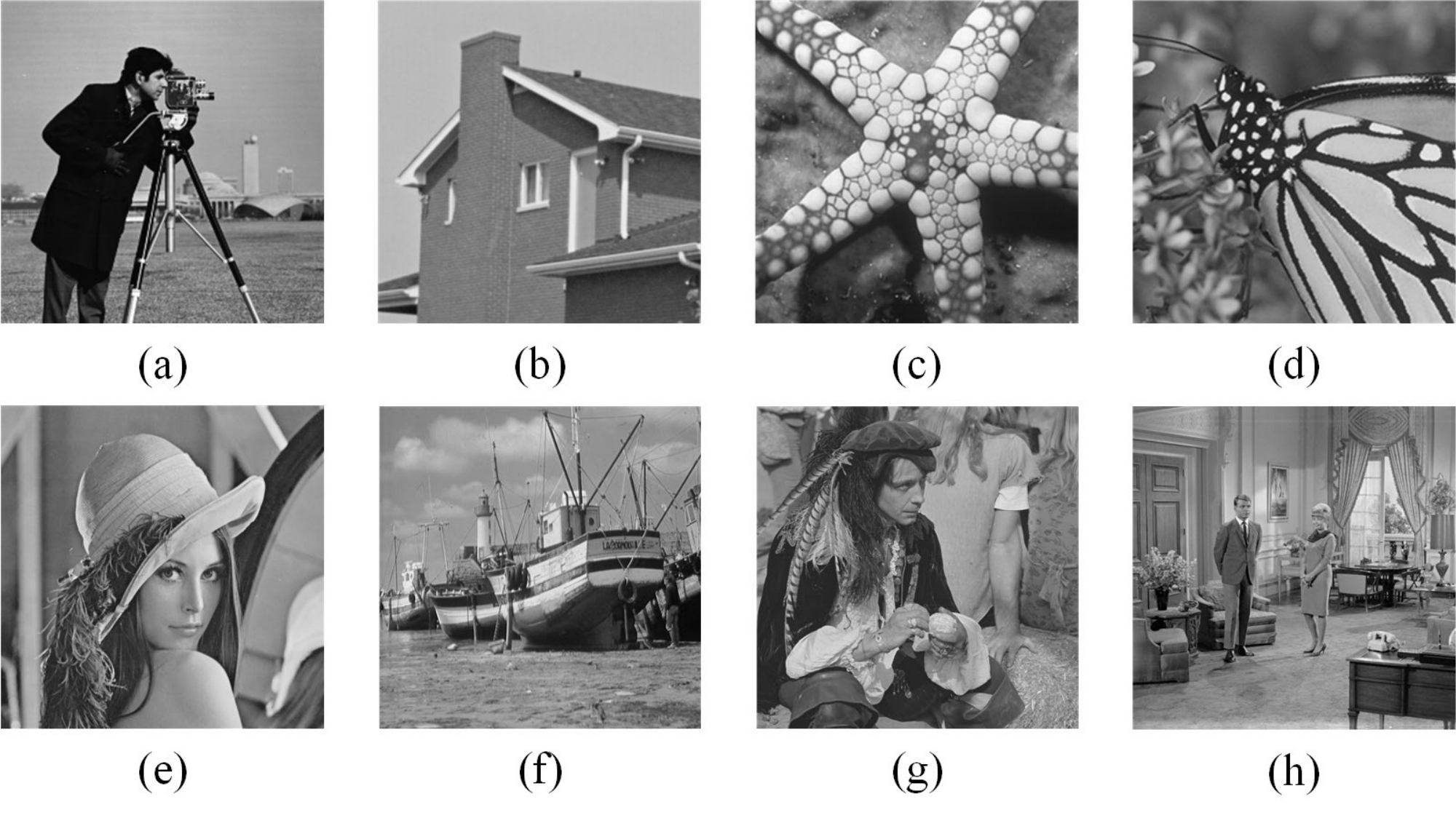}
\caption{The data-set is used for our testing}
\label{Fig3}
\end{figure}

\section{Experimental results}
\begin{figure*}[!ht]
\centering
\includegraphics[width=6.5in]{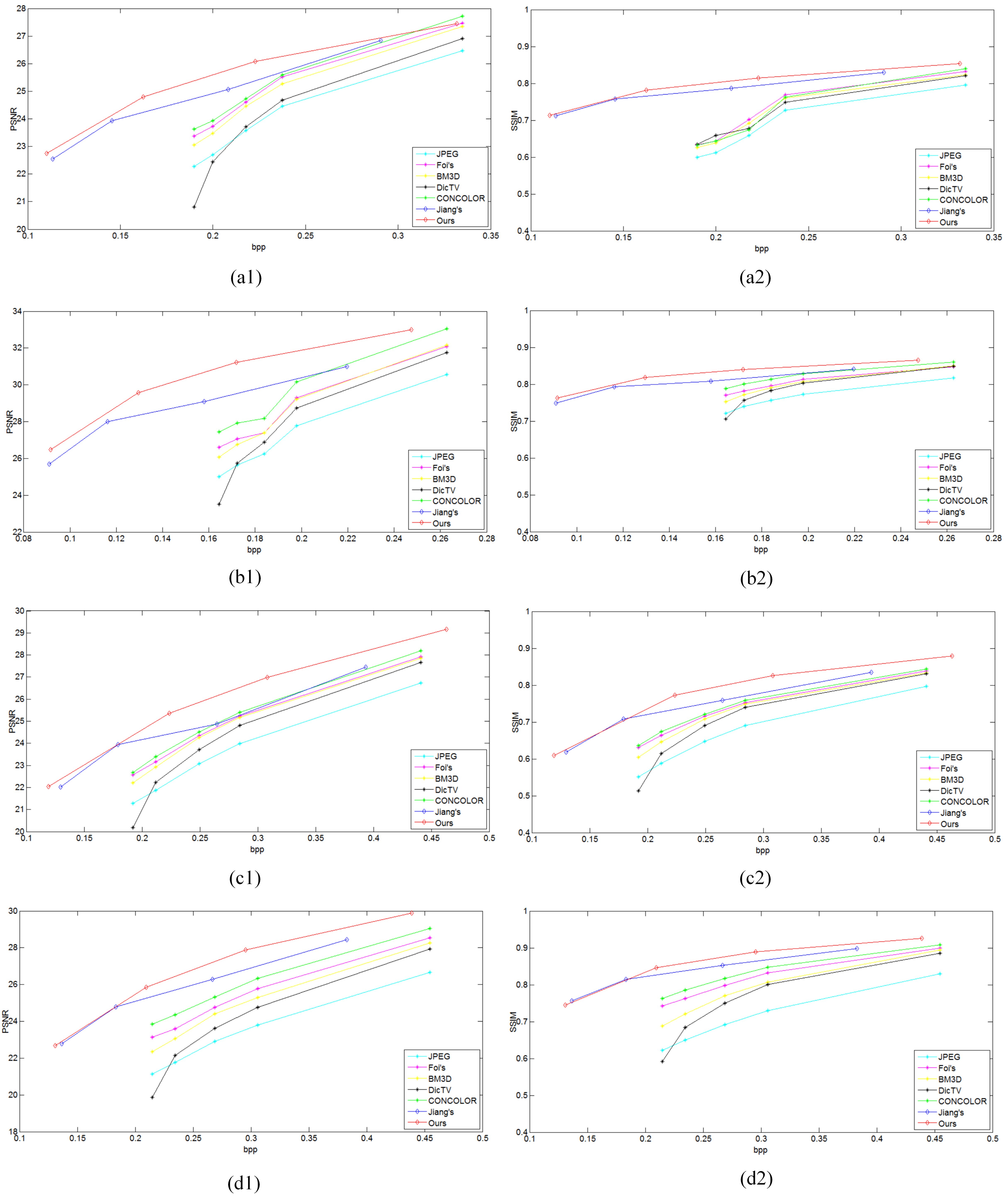}
\caption{The objective measurement comparison on PSNR and SSIM for several state-of-the-art approaches. (a1-a2) are the results of image (a) in Fig. \ref{Fig3}, (b1-b2) are the results of image (b) in Fig. \ref{Fig3}, (c1-c2) are the results of image (c) in Fig. \ref{Fig3}, (d1-d2) are the results of image (d) in Fig. \ref{Fig3}}
\label{Fig4}
\end{figure*}

\begin{figure*}[!ht]
\centering
\includegraphics[width=6.5in]{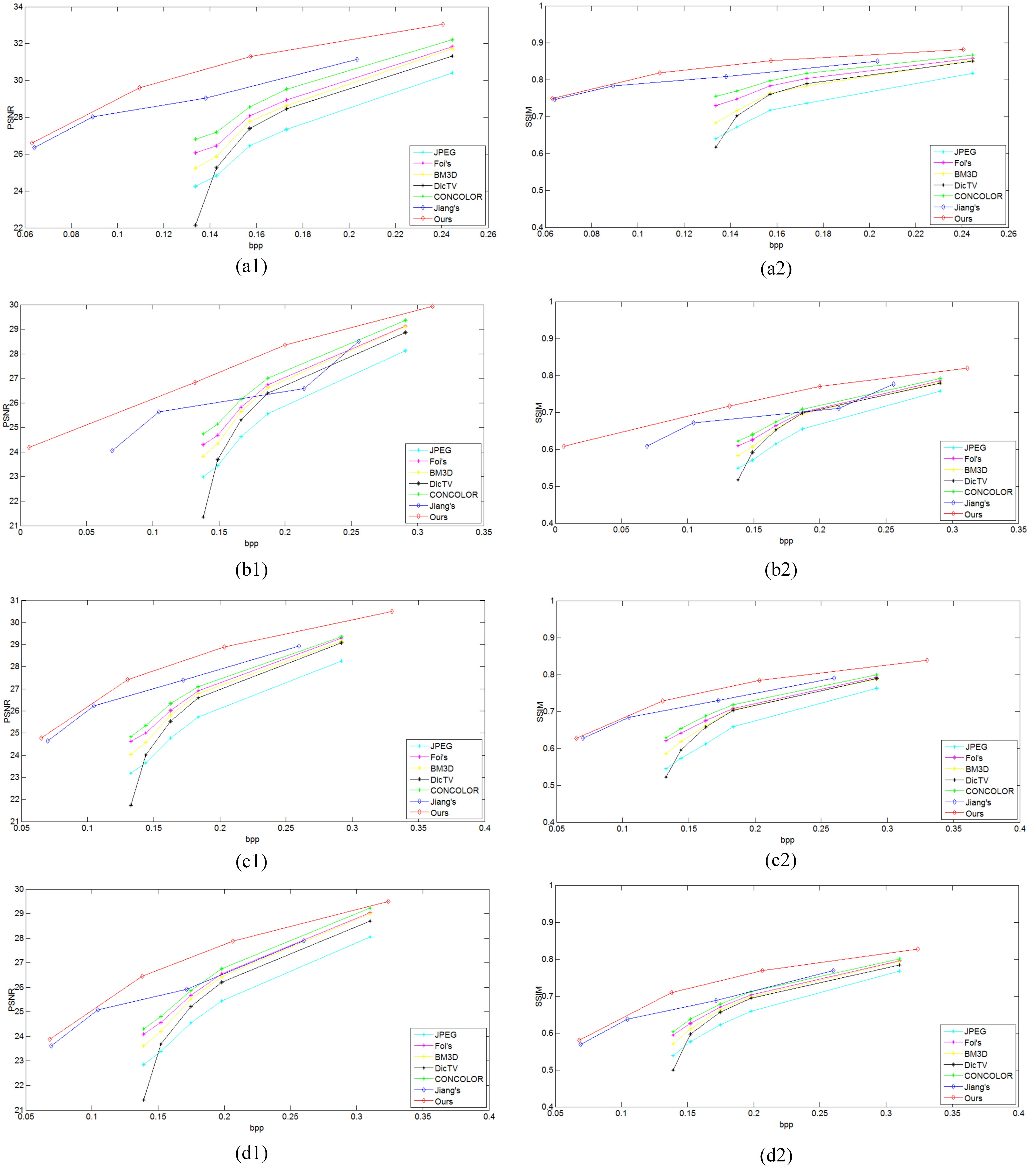}
\caption{The objective measurement comparison on PSNR and SSIM for several state-of-the-art approaches. (a1-a2) are the results of image (e) in Fig. \ref{Fig3}, (b1-b2) are the results of image (f) in Fig. \ref{Fig3}, (c1-c2) are the results of image (g) in Fig. \ref{Fig3}, (d1-d2) are the results of image (h) in Fig. \ref{Fig3}}
\label{Fig5}
\end{figure*}

\begin{figure*}[ht]
\centering
\includegraphics[width=6.3in]{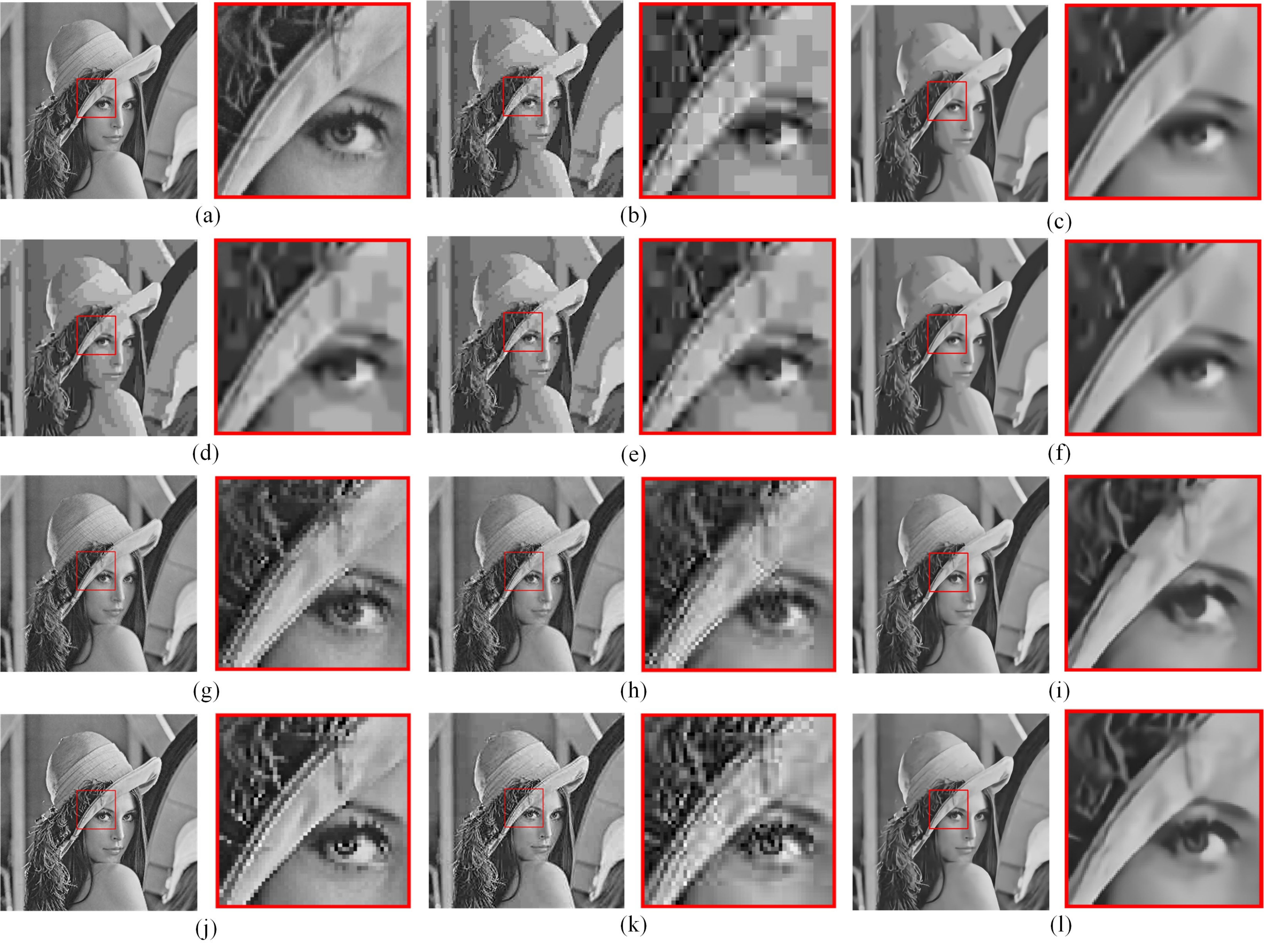}
\caption{The visual comparisons for several state-of-the-art approaches. (a) input image of Lena, (b) compressed image of (a) 26.46/0.718/0.173 (PSNR/SSIM/bpp), (c) Foi's 28.07/0.784/0.173, (d) BM3D 27.76/0.764/0.173, (e) DicTV 27.38/0.761/0.173, (f) CONCOLOR 28.57/0.798/0.173, (g) the output of Jiang's ComCNN, (h) compressed image of (g), (i) the output of Jiang's RecCNN 31.14/0.851/0.204, (j) our FDNN network's output, (k) compressed image of (j), (l) our PPNN network's output 31.31/0.853/0.157 ; Note that the real resolution (g-h) and (j-k) is half of the input image, while all the other images have the same size with (a)}
\label{Fig6}
\end{figure*}

\begin{figure*}[!ht]
\centering
\includegraphics[width=4.5in]{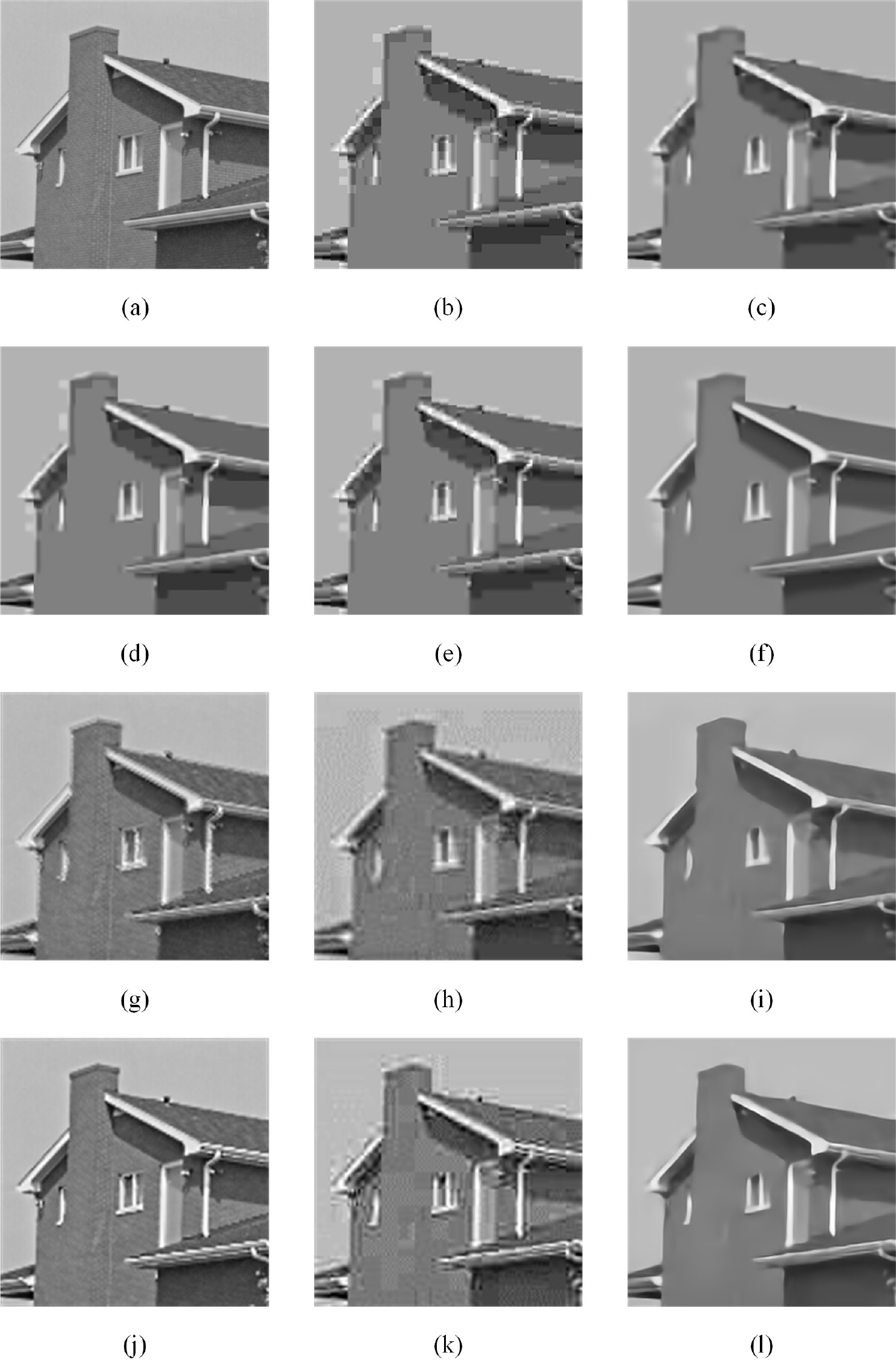}
\caption{The visual comparisons for several state-of-the-art approaches. (a) input image of House, (b) compressed image of (a) 27.77dB/0.773/0.197 (PSNR/SSIM/bpp),  (c) Foi's 29.29dB/0.814/0.197, (d) BM3D 29.21dB/0.808/0.197, (e) DicTV 28.74dB/0.804/0.197, (f) CONCOLOR 30.16dB/0.829/0.197, (g) the output of Jiang's ComCNN, (h) compressed image of (g), (i) the output of Jiang's RecCNN 29.10/0.809/0.157, (j) our FDNN network's output, (k) compressed image of (j), (l) our PPNN network's output 29.57/0.819/0.131; Note that the real resolution (g-h) and (j-k) is half of the input image, while all the other images have the same size with (a)}
\label{Fig7}
\end{figure*}

In order to demonstrate the novelty of the proposed framework, we compare our method with six approaches: JPEG \cite{r1}, Foi's \cite{r11}, BM3D \cite{r12}, DicTV \cite{r15}, CONCOLOR \cite{r18}, and Jiang's \cite{r27}. Here, both Foi's \cite{r11} and BM3D \cite{r12} are the class of image de-noising. The method of Foi's \cite{r11} is specifically designed for de-blocking. The approaches of DicTV \cite{r15}, CONCOLOR \cite{r18} use the dictionary learning or the low-rank model to resolve the problem of de-blocking and de-artifact. Our approach is highly related to the Jiang's approach in \cite{r27}, which is CNN-based methods, so we give many comparisons between them later. %

\subsection{Training details}
Our framework of learning a virtual codec neural network to compress image is implemented with TensorFlow \cite{r35}. The training data-set comes from \cite{r31}, in which 400 images of size 180x180 are included. We augment these data by cropping, rotating and flipping image to build our training data set, in which the total number of image patches with size of 160x160 are 3200 (n=3200). For testing as shown in Fig. \ref{Fig2}, eight images, which are broadly employed for compressed image de-noising or de-artifact, are used to evaluate the efficiency of the proposed method. We train our model using the optimization method of Adam, with the beta1=0.9, beta2=0.999. The initial learning rate of training three convolutional neural network is set to be 0.0001, while the learning rate decays to be half of the initial one once the training step reaches 3/5 of total step. And it decreases to be 1/4 of the initial one when the training step reaches 4/5 of total step. In the \textbf{Algorithm-1}, $K$ equals to 3, $p$ = 60, $q$ is 30, and $m$ is set to be 20.

\subsection{The quality comparison of different methods}
To validate the efficiency of the proposed framework at very low bit-rate, we compare our method with JPEG, Foi's \cite{r11}, BM3D \cite{r12}, DicTV \cite{r15}, CONCOLOR \cite{r18}, and Jiang's \cite{r27}. The JPEG software of image compression in Opencv is used for all the experimental results. The results of Foi's \cite{r11}, BM3D \cite{r12}, DicTV \cite{r15} and CONCOLOR \cite{r18} are got by strictly using the author's open codes with the parameter settings in their papers. However, the highly related method of Jiang's \cite{r27} only give one factor for testing, so we try to re-implement their method with TensorFlow. Meanwhile, to fairly compare with the Jiang's \cite{r27}, we use our FDNN and PPNN to replace its networks of ComCNN and ReCNN for training and testing to avoid the effect of the network's structure design on the experimental results. Additionaly, the training of the Jiang's simulation is achieved according to the framework in \cite{r27}.

It's clear that the Jiang's \cite{r27} has two convolution neural networks, which are directly connected for training to back-propagate the gradient from the ReCNN to the ComCNN, but the proposed method has three convolution neural networks, in which our virtual codec neural network has considered the impacts of codec on the feature description neural network. From this aspect, it can be concluded that the proposed method is superior to the Jiang's \cite{r27}, which doesn't resolve the problem of the codec's effects on the ComCNN's training in Jiang's \cite{r27}. For the comparison of Jiang's \cite{r27} in the following, images are compressed by JPEG with quality factors to be 5, 10, 20, and 40 for their training and testing. Meanwhile, the valid description of input image in the proposed framework is also compressed by JPEG codec with quality factors to be 5, 10, 20, and 40 for training our three convolutional neural networks. Except the proposed method and Jiang's \cite{r27}, four other comparative methods deal with the JPEG compressed image with the full-resolution when the quality factor set is set to be 2, 3, 4, 5, 10. It's worthy to notice that in the proposed framework the JPEG codec is used, but in fact our framework can be applied into most of existing standard codec.

We use the Peak Signal to Noise Ratio (PSNR) and Structural SIMilarity Index (SSIM) as the objective quality's measurement. From the Fig. \ref{Fig4} and Fig. \ref{Fig5}, where bpp denotes the bit-per-pixel, it can be obviously observed that the proposed method has the best objective performance on PSNR and SSIM, as compared to several state-of-the-art approaches: JPEG \cite{r1}, Foi's \cite{r11}, BM3D \cite{r12}, DicTV \cite{r15}, CONCOLOR \cite{r18}, and Jiang's \cite{r27}. Note that the results of Jiang's in Fig. \ref{Fig4} and Fig. \ref{Fig5} are the re-implemented results by us according to \cite{r27}, which has been mentioned previously. Among these methods, CONCOLOR \cite{r18} has a stable objective performance and achieves a great gain in the term of PSNR and SSIM when comparing with Foi's \cite{r11}, BM3D \cite{r12}, and DicTV \cite{r15}. As mentioned above, the proposed method can rightly back-propagate the gradient from the post-processed neural network to the feature description neural network ahead of codec, so our method is nearly optimal as compared with the approach of Jiang's \cite{r27}, which just provides a way to train their two convolutional neural networks together.

We have compared the visual quality of different methods for compression artifact removal, as shown in Fig. \ref{Fig6} and Fig. \ref{Fig7}. From these figures, it can be seen that Foi's \cite{r11} and CONCOLOR \cite{r18} can better remove the block artifacts than BM3D \cite{r12}, DicTV \cite{r15}, but these methods may make image's boundary blurring. Both the proposed method and Jiang's \cite{r27} have better performance on the discontinuity preservation than other methods (Please see the regions of hair and the eyes of Lena in the Fig. \ref{Fig6}), but the proposed method can retain more details and greatly save the bit for compression at the very low bit-rate than Jiang's \cite{r27}.  Meanwhile, we also show the difference between our FDNN's output image and ComCNN's output image in \cite{r27}, which results in the difference of artifact's distributions so that different region may be emphasized and protected during compression, as displayed in the Fig. \ref{Fig6} (g, j) and Fig. \ref{Fig7} (g, j). The artifact's distribution difference, caused by the difference between the description of our FDNN and compact representation of Jiang's ComCNN, leads to the obvious reconstruction differences between Jiang's and ours at the decoder, as displayed in the Fig. \ref{Fig6} (h, k, i, l) and Fig. \ref{Fig7} (h, k, i, l). From the above comparisons, it can be known that the back-propagation of gradient in the feature description network from postprocessing neural network plays a significant role on the effectiveness of feature description and the compression efficiency when combining the neural network with standard codec together to effectively compress image. In a word, the proposed framework provides a good way to resolve the gradient back-propagation problem in the image compression framework with convolutional neural network ahead of a standard codec by learning a virtual codec neural network.

\section{Conclusion}
In this paper, we propose a new image compression framework to resolve the problem of non-differentiability of the quantization function in the lossy image compression by learning a virtual codec neural network at very low bit-rate. Our framework consists of a traditional codec, feature description neural network, post-processing neural network, and virtual codec neural network. Directly learning the whole framework of the proposed method is a intractable problem, so we decompose this challenging optimization problem into three sub-problems learning. Finally, a large number of quantitative and qualitative experimental results have shown the priority of the proposed method than several state-of-the-art methods.




\bibliographystyle{IEEEtran}
\bibliography{IEEEfull,VCODEC}

\begin{thebibliography}{10}
\providecommand{\url}[1]{#1}
\csname url@samestyle\endcsname
\providecommand{\newblock}{\relax}
\providecommand{\bibinfo}[2]{#2}
\providecommand{\BIBentrySTDinterwordspacing}{\spaceskip=0pt\relax}
\providecommand{\BIBentryALTinterwordstretchfactor}{4}
\providecommand{\BIBentryALTinterwordspacing}{\spaceskip=\fontdimen2\font plus
\BIBentryALTinterwordstretchfactor\fontdimen3\font minus
  \fontdimen4\font\relax}
\providecommand{\BIBforeignlanguage}[2]{{%
\expandafter\ifx\csname l@#1\endcsname\relax
\typeout{** WARNING: IEEEtran.bst: No hyphenation pattern has been}%
\typeout{** loaded for the language `#1'. Using the pattern for}%
\typeout{** the default language instead.}%
\else
\language=\csname l@#1\endcsname
\fi
#2}}
\providecommand{\BIBdecl}{\relax}
\BIBdecl

\bibitem{r1}
G.~Wallace, ``{The JPEG still picture compression standard},'' \emph{{IEEE}
  Transactions on Consumer Electronics}, vol.~38, no.~1, pp. xviii--xxxiv,
  1992.

\bibitem{rrr1}
L.~Shen, Z.~Liu, X.~Zhang, W.~Zhao, and Z.~Zhang, ``{An effective CU size
  decision method for HEVC encoders},'' \emph{{IEEE} Transactions on
  Multimedia}, vol.~15, no.~2, pp. 465--470, 2013.

\bibitem{rr1}
J.~Xiong, H.~Li, Q.~Wu, and F.~Meng, ``{A fast HEVC inter CU selection method
  based on pyramid motion divergence},'' \emph{{IEEE} Transactions on
  Multimedia}, vol.~16, no.~2, pp. 559--564, 2014.

\bibitem{r3}
L.~Zhao, A.~Wang, B.~Zeng, and Y.~Wu, ``Candidate value-based boundary
  filtering for compressed depth images,'' \emph{Electronics Letters}, vol.~51,
  no.~3, pp. 224--226, 2015.

\bibitem{r4}
L.~Zhao, H.~Bai, A.~Wang, Y.~Zhao, and B.~Zeng, ``Two-stage filtering of
  compressed depth images with markov random field,'' \emph{Signal Processing:
  Image Communication}, vol.~51, pp. 11--22, 2017.

\bibitem{r5}
L.~Zhao, H.~Bai, A.~Wang, and Y.~Zhao, ``Iterative range-domain weighted filter
  for structural preserving image smoothing and de-noising,'' \emph{Multimedia
  Tools and Applications}, pp. 1--28, 2017.

\bibitem{rr2}
J.~Viéron and C.~Guillemot, ``Real-time constrained tcp-compatible rate
  control for video over the internet,'' \emph{{IEEE} Transactions on
  Multimedia}, vol.~6, no.~4, pp. 634--646, 2004.

\bibitem{rrr2}
S.~Yoo, K.~Choi, and J.~Ra, ``Post-processing for blocking artifact reduction
  based on inter-block correlation,'' \emph{{IEEE} Transactions on Multimedia},
  vol.~16, no.~6, pp. 1536--1548, 2014.

\bibitem{r7}
P.~List, A.~Joch, J.~Lainema, G.~Bjontegaard, and M.~Karczewicz, ``Adaptive
  deblocking filter,'' \emph{{IEEE} Transactions on Circuits and Systems for
  Video Technology}, vol.~13, no.~7, pp. 614--619, 2003.

\bibitem{r6}
H.~Reeve and J.~Lim, ``Reduction of blocking effect in image coding,'' in
  \emph{IEEE International Conference on Acoustics, Speech, and Signal
  Processing}, Boston, Massachusetts, USA, Apr. 1983.

\bibitem{r8}
A.~Liew and H.~Yan, ``Blocking artifacts suppression in block-coded images
  using overcomplete wavelet representation,'' \emph{{IEEE} Transactions on
  Circuits and Systems for Video Technology}, vol.~14, no.~4, pp. 450--461,
  2004.

\bibitem{r9}
N.~Francisco, N.~Rodrigues, E.~Da-Silva, and S.~De-Faria, ``A generic
  post-deblocking filter for block based image compression algorithms,''
  \emph{Signal Processing: Image Communication}, vol.~27, no.~9, pp. 985--997,
  2012.

\bibitem{r10}
C.~Wang, J.~Zhou, and S.~Liu, ``Adaptive non-local means filter for image
  deblocking,'' \emph{Signal Processing: Image Communication}, vol.~28, no.~5,
  pp. 522--530, 2013.

\bibitem{r11}
A.~Foi, V.~Katkovnik, and K.~Egiazarian, ``{Pointwise shape-adaptive DCT for
  high-quality denoising and deblocking of grayscale and color images},''
  \emph{{IEEE} Transactions on Image Processing}, vol.~16, no.~5, pp.
  1395--1411, 2007.

\bibitem{r13}
X.~Zhang, R.~Xiong, X.~Fan, S.~Ma, and W.~Gao, ``Compression artifact reduction
  by overlapped-block transform coefficient estimation with block similarity,''
  \emph{{IEEE} Transactions on Image Processing}, vol.~22, no.~12, pp.
  4613--4626, 2013.

\bibitem{r14}
D.~Sun and W.~Cham, ``{Postprocessing of low bit-rate block DCT coded images
  based on a fields of experts prior},'' \emph{{IEEE} Transactions on Image
  Processing}, vol.~16, no.~11, pp. 2743--2751, 2007.

\bibitem{r15}
H.~Chang, M.~Ng, and T.~Zeng, ``{Reducing artifacts in JPEG decompression via a
  learned dictionary},'' \emph{{IEEE} Transactions on Signal Processing},
  vol.~62, no.~3, pp. 718--728, 2014.

\bibitem{r18}
J.~Zhang, R.~Xiong, C.~Zhao, Y.~Zhang, S.~Ma, and W.~Gao, ``{CONCOLOR:
  Constrained non-convex low-rank model for image deblocking},'' \emph{{IEEE}
  Transactions on Image Processing}, vol.~25, no.~3, pp. 1246--1259, 2016.

\bibitem{r19}
Z.~Wang, D.~Liu, S.~Chang, Q.~Ling, Y.~Yang, and T.~Huang, ``{D3: Deep
  dual-domain based fast restoration of JPEG-compressed images},'' in
  \emph{IEEE Conference on Computer Vision and Pattern Recognition}, Las Vegas,
  NV, United States, Jun. 2016.

\bibitem{r20}
X.~Liu, X.~Wu, J.~Zhou, and D.~Zhao, ``Data-driven soft decoding of compressed
  images in dual transform-pixel domain,'' \emph{{IEEE} Transactions on Image
  Processing}, vol.~25, no.~4, pp. 1649--1659, 2016.

\bibitem{r16}
Y.~Li, F.~Guo, R.~Tan, and M.~Brown, ``{A contrast enhancement framework with
  JPEG artifacts suppression},'' in \emph{European Conference on Computer
  Vision}, Cham, Sep. 2014.

\bibitem{r12}
K.~Dabov, A.~Foi, V.~Katkovnik, and K.~Egiazarian, ``{Image denoising by sparse
  3-D transform-domain collaborative filtering},'' \emph{{IEEE} Transactions on
  Image Processing}, vol.~16, no.~8, pp. 2080--2095, 2007.

\bibitem{r17}
S.~Gu, L.~Zhang, W.~Zuo, and X.~Feng, ``Weighted nuclear norm minimization with
  application to image denoising,'' in \emph{IEEE Conference on Computer Vision
  and Pattern Recognition}, Columbus, OH, USA, Jun. 2014.

\bibitem{rrr3}
D.~Huang, L.~Kang, Y.~Wang, and C.~Lin, ``Self-learning based image
  decomposition with applications to single image denoising,'' \emph{{IEEE}
  Transactions on Multimedia}, vol.~16, no.~1, pp. 83--93, 2013.

\bibitem{r36}
L.~Chen, G.~Papandreou, I.~Kokkinos, K.~Murphy, and A.~Yuille, ``Deeplab:
  semantic image segmentation with deep convolutional nets, atrous convolution,
  and fully connected crfs,'' \emph{{IEEE} Transactions on Pattern Analysis and
  Machine Intelligence}, vol.~PP, no.~99, pp. 1--1, 2016.

\bibitem{r21}
H.~Burger, C.~Schuler, and S.~Harmeling, ``{Image denoising: Can plain neural
  networks compete with BM3D$?$},'' in \emph{IEEE Conference on Computer Vision
  and Pattern Recognition}, Providence, RI USA, Jun. 2012.

\bibitem{r22}
C.~Dong, C.~Loy, K.~He, and X.~Tang, ``Image super-resolution using deep
  convolutional networks,'' \emph{{IEEE} Transactions on Pattern Analysis and
  Machine Intelligence}, vol.~38, no.~2, pp. 295--307, 2016.

\bibitem{r24}
P.~Isola, J.~Zhu, T.~Zhou, and A.~Efros, ``Image-to-image translation with
  conditional adversarial networks,'' in \emph{arXiv: 1611.07004}, 2016.

\bibitem{r23}
L.~Zhao, J.~Liang, H.~Bai, A.~Wang, and Y.Zhao, ``Simultaneously color-depth
  super-resolution with conditional generative adversarial network,'' in
  \emph{arXiv: 1708.09105}, 2017.

\bibitem{r26}
L.~Cavigelli, P.~Hager, and L.~Benini, ``{CAS-CNN: A deep convolutional neural
  network for image compression artifact suppression},'' in \emph{IEEE
  Conference on Neural Networks}, Anchorage, AK, USA, May 2017.

\bibitem{r25}
L.~Galteri, L.~Seidenari, M.~Bertini, and B.~Del, ``Deep generative adversarial
  compression artifact removal,'' in \emph{arXiv: 1704.02518}, 2017.

\bibitem{r28}
G.~Toderici, S.~Malley, S.~Hwang, D.~Vincent, D.~Minnen, S.~Baluja., and
  R.~Sukthankar, ``Variable rate image compression with recurrent neural
  networks,'' in \emph{arXiv: 1511.06085}, 2015.

\bibitem{r29}
J.~Ballé, V.~Laparra, and E.~Simoncelli, ``Variable rate image compression
  with recurrent neural networks,'' in \emph{arXiv: 1611.01704}, 2016.

\bibitem{r30}
M.~Li, W.~Zuo, S.~Gu., D.~Zhao, and D.~Zhang, ``Learning convolutional networks
  for content-weighted image compression,'' in \emph{arXiv: 1703.10553}, 2017.

\bibitem{r27}
F.~Jiang, W.~Tao, S.~Liu, J.~Ren, X.~Guo, and D.~Zhao, ``An end-to-end
  compression framework based on convolutional neural networks,'' \emph{{IEEE}
  Transactions on Circuits and Systems for Video Technology}, 2017.

\bibitem{r33}
M.~Mathieu, C.~Couprie, and Y.~LeCun, ``Deep multi-scale video prediction
  beyond mean square error,'' in \emph{arXiv: 1511.05440}, 2015.

\bibitem{r32}
Z.~Wang, A.~Bovik, H.~Sheikh, and E.~Simoncelli, ``Image quality assessment:
  from error visibility to structural similarity,'' \emph{{IEEE} Transactions
  on Image Processing}, vol.~13, no.~4, pp. 600--612, 2004.

\bibitem{r35}
M.~Abadi, A.~Agarwal, P.~Barham, E.~Brevdo, Z.~Chen, C.~Citro, and et~al.,
  ``Tensorflow: large-scale machine learning on heterogeneous distributed
  systems,'' in \emph{arXiv:1603.04467}, 2016.

\bibitem{r31}
Y.~Chen and T.~Pock, ``Trainable nonlinear reaction diffusion: A flexible
  framework for fast and effective image restoration,'' \emph{{IEEE}
  Transactions on Pattern Analysis and Machine Intelligence}, vol.~39, no.~6,
  pp. 1256--1272, 2017.

\end{thebibliography}
\end{document}